\documentclass[10pt,twocolumn,letterpaper]{article}

\usepackage{cvpr}              

\usepackage{url}
\usepackage{xspace}
\usepackage{booktabs}
\usepackage{makecell}
\usepackage{algorithm}
\usepackage{listings}
\usepackage{amsmath,amsfonts,bm}
\usepackage{algpseudocode}
\usepackage{wrapfig}
\usepackage{enumitem}
\usepackage[table]{xcolor}

\usepackage{multirow}
\usepackage{makecell}
\usepackage{pifont}

\usepackage{graphicx}
\usepackage{caption, subcaption}
\usepackage{multirow}
\usepackage{color}
\definecolor{baselinecolor}{gray}{0.9}
\usepackage{amsmath}
\usepackage{cuted}

\definecolor{cvprblue}{rgb}{0.21,0.49,0.74}
\usepackage[pagebackref,breaklinks,colorlinks,allcolors=cvprblue]{hyperref}

\def\paperID{36603} 
\def\confName{CVPR}
\def\confYear{2026}

\title{PartDiffuser: Part-wise 3D Mesh Generation via Discrete Diffusion}

\author{
    Yichen Yang $^{1\ast}$ \quad
    Hong Li$^{1\ast}$ \quad 
    Haodong Zhu $^{1}$ \quad
    Linlin Yang $^{2\dag}$ \quad \\
    Guojun Lei $^{3}$ \quad
    Sheng Xu $^{1}$ \quad
    Baochang Zhang $^{1}$\\[4pt]
    $^{1}${Beihang University} \quad $^{2}${Communication University of China} \quad $^{3}${Zhejiang University} \
}

\begin{document}

\maketitle

{
    \renewcommand{\thefootnote}{\fnsymbol{footnote}}
    
    \small\footnotetext[1]{Equal contribution. $^{\dag}$Corresponding authors.}
}

\begin{abstract}
Existing autoregressive (AR) methods for generating artist-designed meshes struggle to balance global structural consistency with high-fidelity local details, and are susceptible to error accumulation. 
To address this, we propose PartDiffuser, a novel semi-autoregressive diffusion framework for point-cloud-to-mesh generation. 
The method first performs semantic segmentation on the mesh and then operates in a "part-wise" manner: it employs autoregression between parts to ensure global topology, while utilizing a parallel discrete diffusion process within each semantic part to precisely reconstruct high-frequency geometric features. 
PartDiffuser is based on the DiT architecture and introduces a part-aware cross-attention mechanism, using point clouds as hierarchical geometric conditioning to dynamically control the generation process, thereby effectively decoupling the global and local generation tasks. 
Experiments demonstrate that this method significantly outperforms state-of-the-art (SOTA) models in generating 3D meshes with rich detail, exhibiting exceptional detail representation suitable for real-world applications.
\end{abstract}    
\section{Introduction}

\begin{figure*}[ht]
    \vspace{-12pt}
    \centering
    \includegraphics[width=0.95\textwidth]{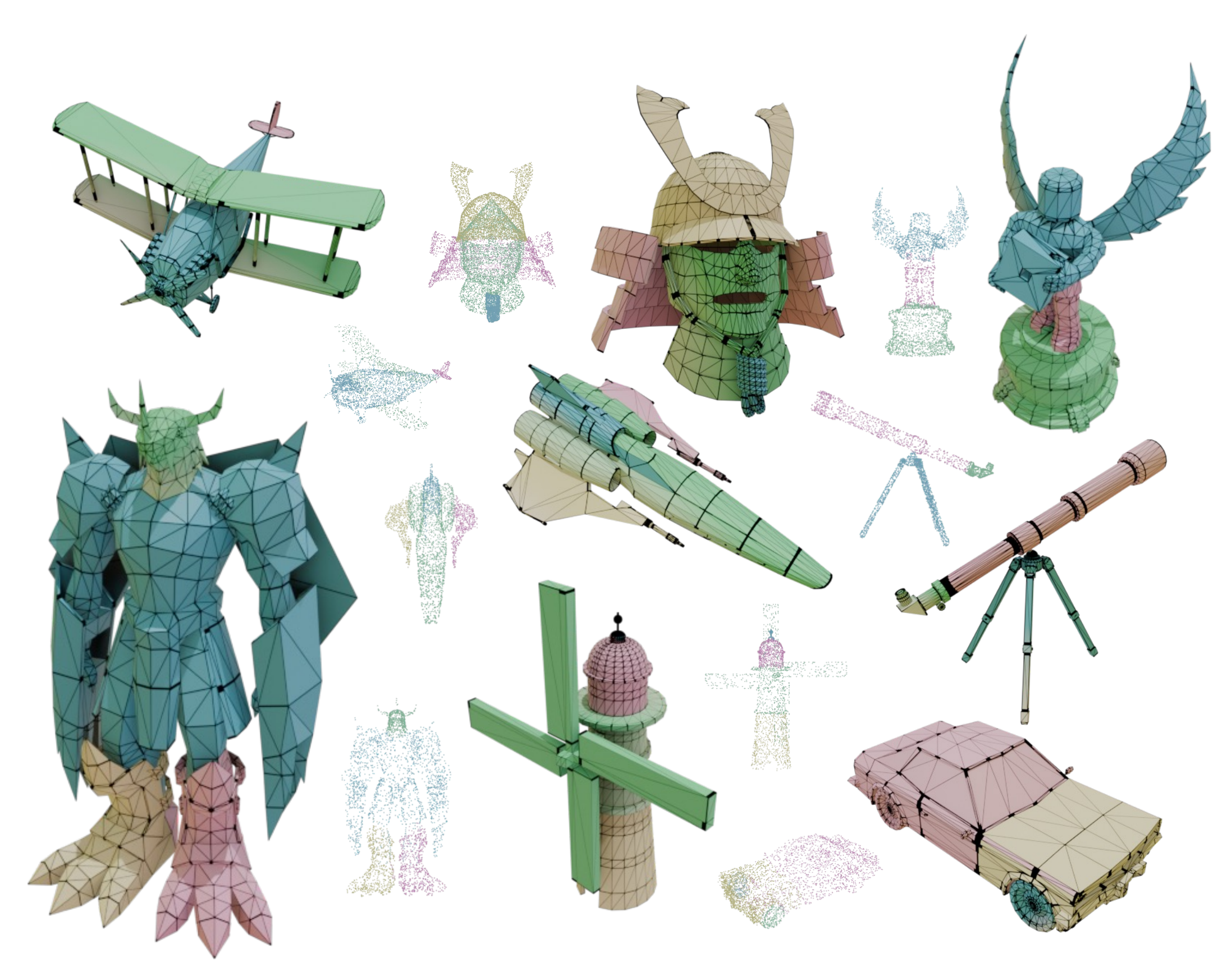}
    \caption{Gallery of our mesh generation results.}
    \label{fig:teaser}
\end{figure*}

The demand for complex, high-fidelity 3D assets is exploding across creative industries, from gaming and VR to film. While various 3D representations exist, triangle meshes remain the dominant format due to their rendering efficiency and flexibility. However, a significant gap persists between meshes generated by automated methods and those meticulously crafted by human artists. Artist-created meshes are distinguished by their clean topology, structural efficiency, and sharp, defining features. In contrast, traditional generative techniques \cite{lorensen1998marching, park2019deepsdf} often yield overly dense, irregular, and unstructured results. This disparity highlights a critical need for generative models that can produce meshes matching the quality and structure of "artist-level" assets.

Inspired by the success of large language models, the dominant paradigm for artist-level mesh generation has shifted to autoregressive (AR) sequence modeling. This approach, seen in seminal works like MeshGPT \cite{siddiqui2024meshgpt} and MeshAnything \cite{chen2024meshanything}, treats a 3D mesh as a one-dimensional sequence of tokens and has achieved significant success in learning structured topology and generating coherent, manifold meshes \cite{wang2024llama, lionar2025treemeshgpt, weng2025scaling, hao2024meshtron, zhao2025deepmesh}. However, this token-by-token serial generation process itself introduces fundamental bottlenecks. First, the long sequential dependency can lead to error accumulation, where a mistake early in the sequence propagates and corrupts the subsequent generation. Second, the model is forced to trade off between global structural coherence and high-fidelity local details. To prioritize global topological correctness and consistency, the model tends to over-smooth or simplify fine-grained, high-frequency geometric features. This inherent "global vs. local" conflict, compounded by the risk of error accumulation, remains a key challenge for current SOTA methods.

To address these challenges, our core insight is to decouple these two tasks: Global structure can be guaranteed through "part-level" dependencies, while local details can be finely modeled "within-part". This semi-autoregressive sampling strategy has recently shown compelling results in diffusion language models \cite{arriola2025block,nie2025llada,ye2025dream}, as it effectively addresses this trade-off. We propose adapting this strategy: employ autoregression between parts to maintain global structure, while using a parallel diffusion model \cite{austin2021structured} within parts to focus on high-fidelity local details. This "Part-Diffusion" process has the potential to overcome the aforementioned trade-offs.

Thus, we propose \textbf{PartDiffuser}, a novel framework for 3D mesh generation. PartDiffuser operates in a "part-wise" manner, generating one complete semantic part at a time. For each part, the model utilizes a DiT-based \cite{peebles2023dit} discrete diffusion process to generate all tokens for that part in parallel. This design allows the model to fully concentrate on the local details of each part during its generation (addressing the local-fidelity issue), while the semi-autoregressive process ensures all parts are assembled correctly (maintaining global consistency). Crucially, by limiting autoregressive steps to the part level rather than the token level, this framework also mitigates the long-range error accumulation inherent in purely sequential models. This achieves both global consistency and local high-fidelity.

The main contributions of this paper are summarized as follows:
\begin{itemize}
    \item We propose PartDiffuser, a novel semi-autoregressive diffusion framework that resolves the conflict between global structure and local details in 3D mesh generation.
    \item We introduce a part-aware cross-attention mechanism that dynamically guides the generation process with hierarchical geometric features, enhancing local fidelity.
    \item We conducted extensive experiments on various datasets. The results demonstrate that our method significantly outperforms SOTA methods in the reconstruction of high-frequency details.
\end{itemize}
\section{Related Works}

\subsection{Artist Mesh Generation}

Triangle meshes are the de facto standard for 3D assets in fields like gaming and virtual reality, prized for their efficiency in rendering and manipulation. However, traditional generation methods, which convert implicit fields or volumetric grids \cite{park2019deepsdf, mildenhall2021nerf,long2024wonder3d,xiang2025structured} to meshes via algorithms like Marching Cubes \cite{lorensen1998marching}, often produce overly dense and irregular topologies. These results lack the optimized structure and sharp features of meshes created by human artists. To address this gap, a dedicated line of research has focused on generating "artist-created" meshes directly. Inspired by large language models, the dominant paradigm in this area has become autoregressive sequence modeling. Seminal works like MeshGPT \cite{siddiqui2024meshgpt}, followed by MeshAnything \cite{chen2024meshanything}, LLaMA-Mesh \cite{wang2024llama}, MeshXL \cite{chen2024meshxl}, Meshtron \cite{hao2024meshtron}, treat the mesh as a 1D sequence of tokens (representing vertices and faces) and have shown success in learning structured topology to generate coherent, manifold assets, often conditioned on inputs like point clouds or images.

Despite this success, the autoregressive, token-by-token process presents a fundamental bottleneck in achieving high-fidelity local detail. The sequential dependency struggles to model long-range relationships and can cause error accumulation, forcing a trade-off between global structural coherence and high-frequency local features. While many efforts have focused on improved tokenization to manage sequence length \cite{weng2025scaling,chen2024meshanything,lionar2025treemeshgpt}, and others like DeepMesh \cite{zhao2025deepmesh} have begun to apply subsequent reinforcement learning to improve outputs, the underlying sequential nature still often causes fine-grained geometric details to be over-smoothed or simplified as the model prioritizes topological consistency. Thus, a significant open challenge remains in developing a framework that can efficiently generate rich, high-fidelity local details, unconstrained by the trade-offs inherent in a global, sequential generation process.

\subsection{Part-level 3D Generation}

The fundamental concept behind generating shapes at the part-level is the decomposition of intricate objects into meaningful semantic units. This approach supplies extensive prior knowledge for structure-aware rebuilding and controllable creation, aligning well with the part-based workflow of human artists. Some studies relied on a two-step process: explicit segmentation followed by reconstruction. For instance, PartGen \cite{chen2025partgen} tackles 3D object decomposition by using multi-view segmentation, after which it individually completes and reconstructs each component in 3D. PhyCAGE \cite{yan2024phycage} takes an alternative route, employing physical regularization techniques to break down non-rigid objects. Other methods work directly from partial segmentation data; HoloPart \cite{yang2025holopart} completes the geometry of parts based on initial 3D segmentation results , while Part123 \cite{liu2024part123} can reconstruct a shape from a single image, concurrently forecasting its semantic components and their spatial configuration.

Another category of methods leverages Diffusion Transformer (DiT) models to achieve part-level generation, often bypassing the need for explicit segmentation results. Both PartCrafter \cite{lin2025partcrafter} and PartPacker \cite{tang2025partpacker}, for example, utilize multi-instance DiTs to automatically produce components. PartCrafter can operate from a single image to jointly form multiple, semantically distinct parts, enabling an end-to-end, part-conscious 3D mesh synthesis. PartPacker extends this concept by introducing a dual-volume DiT, which improves efficiency by modeling complementary spatial volumes. Frankenstein \cite{yan2024frankenstein} employs a comparable strategy, using a VAE to consolidate multiple SDFs within a latent triplane space. Additionally, AutoPartGen \cite{chen2025autopartgen} uses a latent diffusion model to sequentially generate parts in an autoregressive fashion, though this method is noted for being computationally intensive and offering limited user control. Furthermore, BANG \cite{zhang2025bang} dynamically divides complete 3D assets into individual parts through generative exploded dynamics. However, these methods often rely on continuous latent spaces or intermediate representations, leading to inevitable information loss during conversion to final meshes and difficulty in maintaining precise topological consistency across generated parts.

\subsection{Discrete Diffusion Models} 

Diffusion models have emerged as a powerful generative paradigm, demonstrating state-of-the-art performance, particularly in continuous domains like image synthesis \cite{ho2020denoising,dhariwal2021diffusion,ho2022classifier,rombach2022high}. Inspired by their success in continuous domains, recent efforts to adapt diffusion models to discrete data, such as text generation \cite{li2022diffusion, austin2021structured}, have also shown significant progress.

The dominant approach currently is the Masked Discrete Diffusion model \cite{austin2021structured}. Its generation process begins with a fully masked sequence. In each step, the model predicts all tokens in parallel, after which a scheduling policy unmasks high-confidence predictions while remasking uncertain ones for subsequent refinement \cite{khanna2025mercury, wu2025fastdllm}. This paradigm has been scaled to large models like DiffuLLaMA \cite{gong2024scaling}, LLaDA \cite{nie2025llada}, and Dream \cite{ye2025dream}, demonstrating performance competitive with strong AR counterparts. This approach has also been successfully applied to various downstream tasks, including code generation \cite{gong2025diffucoder} and computational biology for protein design \cite{wang2024dplm, goel2024memdlm}. In the 3D mesh domain, TSSR \cite{song2025topology} addresses topological accuracy in fully parallel discrete diffusion by decoupling the process into topology sculpting and shape refining stages.

Rather than mitigating the challenges of a fully parallel approach, Hybrid AR-Diffusion models have been introduced. A representative strategy is the block-wise semi-autoregressive framework, such as BD3-LM \cite{arriola2025block}. This method generates large blocks of tokens autoregressively, but within each block, it generates all tokens in parallel using diffusion. This hybrid, block-wise paradigm presents a compelling solution for 3D mesh generation, as it addresses the trade-off between global structural coherence and high-frequency local detail.

\section{Method}

\begin{figure*}[ht]
    \centering
    \includegraphics[width=0.98\textwidth]{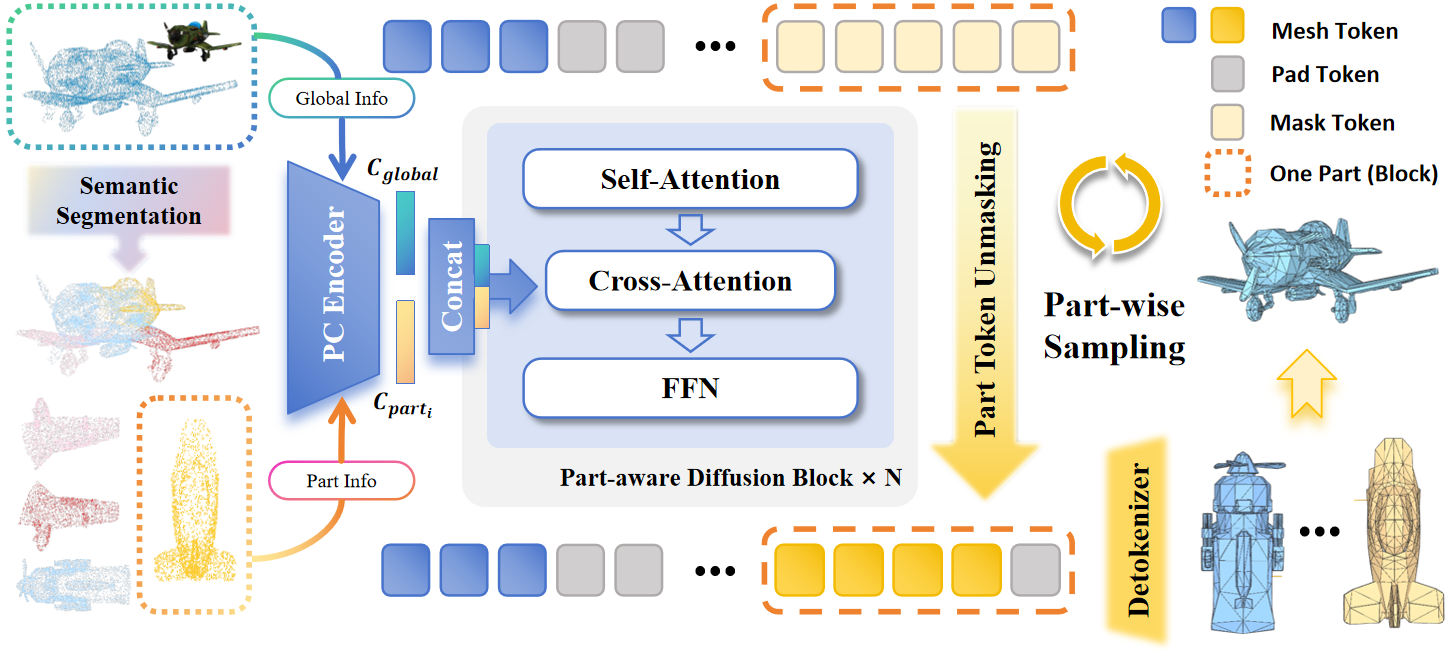}
    \caption{An overview of our PartDiffuser framework. The process begins with semantic segmentation of the input point-cloud using PartField \cite{liu2025partfield}. A pre-trained point cloud encoder, Michelangelo \cite{zhao2023michelangelo}, extracts hierarchical geometric conditions. These conditions are dynamically injected via cross-attention into the Part-aware Diffusion Blocks, which guides the semi-autoregressive "Part-wise Sampling" process of our Discrete Diffusion Model. to generate the final mesh.}
    \label{fig:pipeline}
\end{figure*}

\subsection{Overall Framework}

We propose a novel framework for 3D mesh generation that operates at a semantic part level, based on a DiT-based \cite{peebles2023dit} discrete diffusion model augmented with a Cross-Attention mechanism designed to integrate precise geometric guidance—derived from a point cloud as a single global feature and distinct local part features—into the denoising process. To align the model with this structure, we first preprocess the mesh by segmenting it and serializing each part into a corresponding token block (Sec.~\ref{segmentation}). This design enables our key generation strategy (Sec.~\ref{generation}): a semi-autoregressive sampling process where the model applies a full denoising schedule for each part's generation. Simultaneously, its Cross-Attention layers are dynamically configured to attend only to the global feature and the specific part feature, ensuring precise, part-by-part geometric control. The overall process is illustrated in Figure \ref{fig:pipeline}.

\subsection{Segmented Mesh Representation}
\label{segmentation}

\subsubsection{Semantic Segmentation}

To fully leverage the capability of the discrete diffusion model for learning local bidirectional context and dependencies, we first perform semantic segmentation on the global mesh. This strategy ensures that each resulting mesh part possesses greater internal consistency. We employ PartField \cite{liu2025partfield} to accomplish this semantic segmentation, and the target number of clusters specified for PartField is randomly selected within a predefined range. The sampled point cloud adopts the same segmentation as the mesh, used later for geometric conditioning.

After segmentation, we construct an adjacency matrix derived from the neighborhood relationships of these parts. To increase dataset diversity and model robustness, we randomly select a vertex to initiate a Breadth-First Search traversal (BFS). By exploring neighbors layer by layer, this BFS traversal naturally preserves locality, ensuring that parts which are spatially close on the adjacency graph will also be placed relatively close to each other in the resulting 1D sequence. This preservation of locality is critical for learning local context, as it makes it easier for the model to capture and model the strong dependencies between physically adjacent parts.

\subsubsection{Mesh Tokenization}

To adapt the mesh data for the discrete diffusion model, it must first be processed into a serialized sequence of discrete tokens and padded to a uniform length. To achieve this, we adopt the Blocked and Patchified Tokenization (BPT) method from \cite{weng2025scaling}. BPT addresses redundancy by grouping adjacent faces into patches, each represented by a central vertex followed by its neighbors. It avoids special break tokens by using a dual-block vocabulary, where a unique token type for the center vertex implicitly signals the start of a new patch. We use padding to fill the encoded sequence of each part to a uniform length and then assemble them.

\subsection{Diffusion for Part-wise Mesh Generation}
\label{generation}

\subsubsection{Hierarchical Geometric Conditioning}

We propose a hierarchical geometric guidance mechanism to condition the diffusion process. A pre-trained point cloud encoder, Michelangelo \cite{zhao2023michelangelo}, is employed to derive a structured conditioning representation, $C_{\text{pc}}$, from the input point cloud. This representation is composed of two distinct levels of detail: first, a single global feature vector, $C_{\text{global}}$, which captures the holistic shape of the entire mesh. Second, a sequence of $N$ part-specific feature vectors, $\{C_{\text{part}_1}, \dots, C_{\text{part}_N}\}$, where each $C_{\text{part}_i}$ encapsulates the local geometry for the $i$-th semantic part. The final representation $C_{\text{pc}}$ is formed by concatenating the global feature vector with the sequence of $N$ part feature vectors, directly linking the geometric part decomposition to the generative output structure.

\subsubsection{Part-Aware Diffusion Block}
\label{diffusion_block}

We modify the standard DiT block \cite{peebles2023dit} by inserting a cross-attention module. This Part-Aware Diffusion Block is specifically designed to dynamically inject the hierarchical geometric conditions. For this new layer, the Query ($Q$) is the sequence of intermediate noisy mesh tokens, $Z$, passed from the preceding self-attention module. The Key ($K$) and Value ($V$) are dynamically constructed from the hierarchical conditioning tensor $C_{\text{pc}}$, which is composed of a global feature $C_{\text{global}}$ and $N$ part-specific features $\{C_{\text{part}_1}, \dots, C_{\text{part}_N}\}$.

The core of our method lies in the dynamic behavior of this block, which adapts to enforce a precise correspondence between token blocks and their specific part geometries based on the operational mode. Let the intermediate token sequence before self-attention be $Z \in \mathbb{R}^{B \times L \times D}$. The final output $Z_{\text{out}}$ is computed as:
$$Z^{\prime} = \text{SelfAttn}(\text{LN}(Z)) + Z$$
$$\hat{Z} = \text{CrossAttn}(Q=\text{LN}(Z^{\prime}), K=C_{\text{dyn}}, V=C_{\text{dyn}}) + Z^{\prime}$$
where $C_{\text{dyn}}$ is the dynamically selected conditioning feature, constructed by concatenating $C_{\text{global}}$ with $C_{\text{part}_i}$ corresponding to the semantic part being processed. Finally, $\hat{Z}$ is processed by a standard Feed-Forward Network with a residual connection to produce the ultimate block output $Z_{\text{out}}$.

The self-attention module employs a specialized composite mask as an important tool to achieve part-wise generation. Building upon block-wise semi-autoregressive principles \cite{arriola2025block}, it permits full, bidirectional attention within each token block, while enforcing an autoregressive attention between token blocks, allowing part $i$ to attend only to previous parts $X_{<i}$. Furthermore, a Block-Aware Padding Mask works in coordination to manage padding token visibility across these modes. We defer the detailed visualization and formulation of this composite mask to the Appendix.

\subsubsection{Model Objective}

The entire generation process is formalized as a semi-autoregressive conditional diffusion model. The overall likelihood of the mesh $X$, composed of $N$ parts $\{X_1, \dots, X_N\}$, is factorized semi-autoregressively based on the geometric condition $C_{\text{pc}}$ and previously generated parts $X_{<i}$:
$$p_{\theta}(X | C_{\text{pc}}) = \prod_{i=1}^{N} p_{\theta}(X_i | X_{<i}, C_{\text{pc}})$$
Each part-specific distribution $p_{\theta}(X_i | X_{<i}, C_{\text{pc}})$ is modeled using a discrete diffusion process. The model $p_{\theta}$ is trained to predict the clean data $X_i^0$ (the original part tokens) given a noisy version $X_i^t$ at timestep $t$, the conditioning from previous parts $X_{<i}$ (handled by self-attention), and the specific geometric guidance $C_{\text{dyn}, i} = [C_{\text{global}}, C_{\text{part}_i}]$ (handled by cross-attention). 

Following the simplified objective for masked diffusion, the training objective $\mathcal{L}$ for the $i$-th part is:
$$\mathcal{L}_i = \mathbb{E}_{t, X_i^t \sim q(\cdot | X_i^0)} \left[ w(t) \left( - \log p_{\theta}(X_i^0 | X_i^t, X_{<i}, C_{\text{dyn}, i}) \right) \right]$$
where $q$ is the forward noise process and $w(t)$ is a time-dependent loss weight derived from the noise schedule. The dynamic context $C_{\text{dyn}}$ in the equations above is the key mechanism for part-aware control.

\subsubsection{Training and Sampling Strategy}
\label{training strategy}
Following BD-LM \cite{arriola2025block}, our model operates in a parallel mode during training, receiving the complete token sequence logically divided into noisy and clean blocks to efficiently teach the association between token blocks and part-level geometric features. For every token block, the cross-attention layers are conditioned on its specific geometric feature $[C_{\text{global}}, C_{\text{part}_i}]$. 

During inference, the generation process is semi-autoregressive, reconstructing one semantic part at a time over $N$ strides. When generating the $i$-th part, we employ the sampling strategy from LLaDA \cite{nie2025llada} fro intra-block generation, iteratively predicting and selectively remasking tokens. The cross-attention condition is dynamically set to $[C_{\text{global}}, C_{\text{part}_i}]$, guiding the denoising process with both global and local geometries. Once fully denoised, the condition updates for the subsequent part, repeating until the entire mesh is generated.

\section{Experiments}
\label{sec:experiments}

\begin{figure*}[h]
    \centering
    \includegraphics[width=0.9\textwidth]{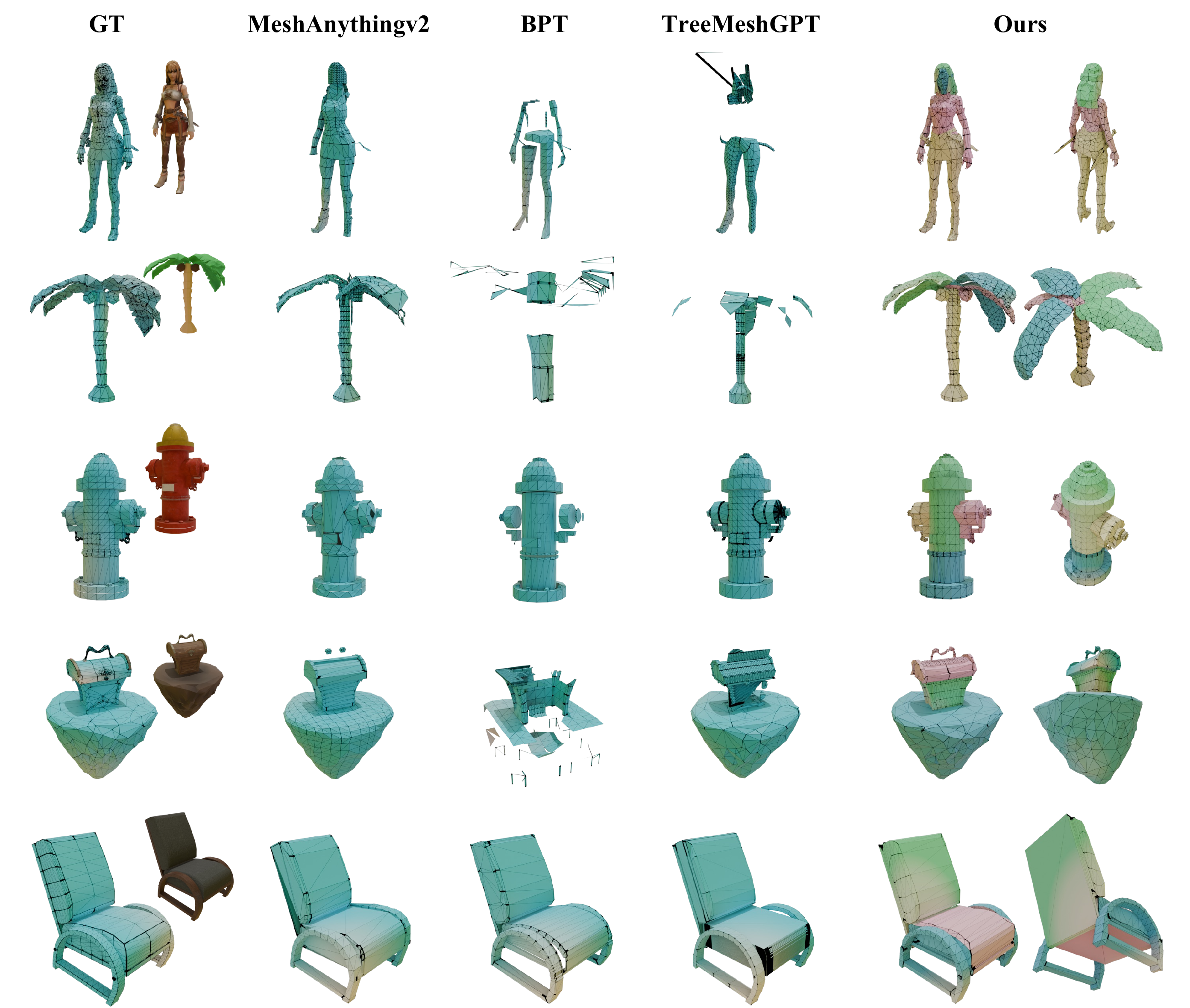}
    \caption{ Visual comparison of PartDiffuser with Baselines.}
    \label{fig:visual}
\end{figure*}

\subsection{Implementation Details}

The training data is curated from a filtered combination of the Objaverse\cite{deitke2023objaverse} and 3DFront\cite{fu20213dfront} datasets. We first process the meshes using PartField\cite{liu2025partfield} to obtain semantic segmentations. A critical preprocessing step involves tokenizing each semantic part, and we apply a strict filtering process: we discard any samples where any single segmented part results in an encoded sequence exceeding 1024 tokens. This filtering ensures that individual parts remain within our designated block size. Our final curated dataset comprises approximately 81K samples.

Our model is built upon a discrete diffusion framework similar to BD-LM\cite{arriola2025block}, which we adapt to incorporate the proposed hierarchical point cloud features \ref{generation}. The model comprises 0.3B parameters. We constrain the maximum sequence length to 4096 tokens, and the semi-autoregressive block size is set to 1024 tokens, aligning with our dataset's filtering criteria. The training process was divided into two stages: pre-training on 8 NVIDIA H100 80GB GPUs for 3 days, followed by fine-tuning on 4 NVIDIA H100 80GB GPUs for two weeks.

\subsection{Metrics}

To quantitatively evaluate the fidelity of our generated meshes, we measure the geometric similarity between our outputs and their corresponding ground-truth shapes using a standard suite of metrics widely adopted for 3D reconstruction. Specifically, we report the Chamfer Distance (CD), which computes the average squared distance between nearest neighbors for point clouds sampled from both surfaces, and the Hausdorff Distance (HD), which measures the maximum distance between the two surfaces. We also utilize the Earth Mover's Distance (EMD), which calculates the minimum cost to transform one point distribution into the other, capturing overall shape similarity. Finally, to evaluate the balance of reconstruction accuracy and completeness, we also report the F1-Score (F1).

\subsection{Point Cloud to Mesh Generation}
\subsubsection{Baselines}
\label{baselines}

We conduct comparative experiments under point-cloud conditioning against three open-source, state-of-the-art mesh generation methods: MeshAnythingV2\cite{chen2025meshanythingv2}, BPT\cite{weng2025scaling}, and TreeMeshGPT\cite{lionar2025treemeshgpt}. The comparison includes both quantitative measurements and qualitative visualization of the results. The test set consists of 300 samples randomly selected from each of Objaverse\cite{deitke2023objaverse}, HSSD\cite{khanna2024hssd}, and 3DFront\cite{fu20213dfront} for the experiment.

\subsubsection{Comparison Results}

Table \ref{tab:comparison_results} presents the quantitative comparison of PartDiffuser against the baselines across the three test datasets. 
On the HSSD and 3DFront datasets, which primarily consist of regular, furniture-style objects, our method demonstrates highly competitive performance. For 3DFront, our model ranks first across three of the four metrics, specifically CD, HD, and EMD. It also achieves a close second on F1-Score with a value of 0.453, just behind TreeMeshGPT's 0.462. On HSSD, it secures the best performance for EMD at 0.059 and F1-Score at 0.471, while remaining highly competitive with TreeMeshGPT on CD and HD.

The advantage of PartDiffuser becomes most apparent on the more complex and diverse Objaverse dataset. Here, our model significantly outperforms all other methods across every single metric. Notably, our CD of 17.813 marks a 27\% improvement over the next-best result from MeshAnythingV2. Furthermore, our F1-Score of 0.343 is nearly 20\% higher than the second-best score of 0.285. This demonstrates our model's superior capability in reconstructing high-fidelity geometry, especially when handling complex and varied shapes, while maintaining strong performance on more structured objects.

Figure \ref{fig:visual} presents the qualitative visualization results on the Objaverse dataset. The generations from BPT appear chaotic, producing few valid or coherent meshes. This suggests that the model struggles significantly with the complexity and diversity of the dataset, failing to learn a stable representation for structured topology. In contrast, MeshAnythingV2 and TreeMeshGPT demonstrate a better capacity to handle the overall global structure in most cases. However, their generation results frequently suffer from localized artifacts, such as overly complex or excessively simplified patches of the mesh. Furthermore, their autoregressive process can lead to error accumulation, causing topological errors to propagate and resulting in generated meshes that deviate significantly from the input.

\begin{table}[t]
\centering
\caption{Quantitative comparison for point cloud to mesh generation against state-of-the-art baselines. Datasets include Objaverse\cite{deitke2023objaverse}, HSSD\cite{khanna2024hssd}, and 3DFront\cite{fu20213dfront}. The Chamfer Distance (CD) metric is scaled by $10^3$. The \textbf{best} results are highlighted in bold, and the \underline{second-best} results are underlined.}
\label{tab:comparison_results}
\footnotesize
\setlength{\tabcolsep}{4pt}
\begin{tabular}{llcccc}
\toprule
Dataset & Method & CD $\times 10^3 \downarrow$ & HD $\downarrow$ & EMD $\downarrow$ & F1 $\uparrow$ \\
\midrule
\multirow{4}{*}{Objaverse} & MeshAnythingV2 & \underline{24.402} & \underline{0.253} & \underline{0.136} & \underline{0.285} \\
                         & BPT            & 86.837 & 0.493 & 0.256 & 0.138 \\
                         & TreeMeshGPT    & 36.938 & 0.306 & 0.157 & 0.279 \\
                         & Ours           & \textbf{17.813} & \textbf{0.238} & \textbf{0.115} & \textbf{0.343} \\
\midrule
\multirow{4}{*}{HSSD}      & MeshAnythingV2 & 9.197  & 0.170 & 0.077 & 0.383 \\
                         & BPT            & 14.029 & 0.186 & 0.093 & 0.369 \\
                         & TreeMeshGPT    & \textbf{4.522}  & \textbf{0.135} & \underline{0.061} & \underline{0.441} \\
                         & Ours           & \underline{5.364}  & \underline{0.138} & \textbf{0.059} & \textbf{0.471} \\
\midrule
\multirow{4}{*}{3DFront}   & MeshAnythingV2 & 10.406 & 0.169 & 0.076 & 0.420 \\
                         & BPT            & 15.652 & 0.203 & 0.096 & 0.421 \\
                         & TreeMeshGPT    & \underline{7.793}  & \underline{0.161} & \underline{0.073} & \textbf{0.462} \\
                         & Ours           & \textbf{6.461}  & \textbf{0.147} & \textbf{0.068} & \underline{0.453} \\
\bottomrule
\end{tabular}
\end{table}

In contrast, our PartDiffuser demonstrates a superior capability in reconstructing high-fidelity geometry. By decoupling the generation process, our method successfully maintains global structural integrity through part-level autoregression, while simultaneously modeling fine-grained local details using parallel discrete diffusion within each part. As seen in the examples, PartDiffuser effectively captures intricate local structures and complex surface features. Moreover, due to the part-wise generation strategy, PartDiffuser avoids producing overly complexified or simplified local meshes.

\subsection{Ablation Study}
\label{sec:ablation}

We conduct an ablation study to validate the effectiveness of the components within our proposed hierarchical geometric conditioning mechanism. Our full model, denoted as \textbf{Ours (Full)}, utilizes both a single global shape feature $C_{global}$ and $N$ part-specific local features $\{C_{part_{i}}\}$, which are dynamically integrated via cross-attention. We compare this approach against two primary variants. First, \textbf{w/ Global only}, which uses only the global feature $C_{global}$ as the condition for all token blocks, removing the part-specific features. Second, \textbf{w/ Parts only}, which uses only the local features $\{C_{part_{i}}\}$ while removing the $C_{global}$ context.
    
Training each variant from scratch would be computationally expensive. Therefore, to ensure a fair comparison while managing resources, we initialize both the 'w/ Global only' and 'w/ Parts only' models with the pre-trained weights of our full model. We then modify their cross-attention mechanisms to accept only the ablated conditioning input. Following this modification, we continue training the entire model for 30\% of the original training steps, allowing all parameters to fully adapt to the new conditioning structure. Results are reported on the Objaverse\cite{deitke2023objaverse} test set in Table \ref{tab:ablation_conditioning}.

\begin{table}[t]
\centering
\caption{Ablation study on the components of our hierarchical geometric conditioning mechanism. All models are evaluated on the test set from Objaverse\cite{deitke2023objaverse}.}
\label{tab:ablation_conditioning}
\footnotesize
\begin{tabular}{lcccc}
\toprule
Method & CD $\times 10^3 \downarrow$ & HD $\downarrow$ & EMD $\downarrow$ & F1 $\uparrow$ \\
\midrule
\textbf{Ours (Full)} & \textbf{17.813} & \textbf{0.238} & \textbf{0.115} & \textbf{0.343} \\
w/ Global only   & \underline{29.125} & \underline{0.297} & \underline{0.144} & \underline{0.281} \\
w/ Parts only    & 54.728 & 0.356 & 0.205 & 0.239 \\
\bottomrule
\end{tabular}
\end{table}

As shown in Table \ref{tab:ablation_conditioning}, the results on the Objaverse test set demonstrate that our full model, \textbf{Ours (Full)}, achieves the best performance across these metrics. This superior performance validates the effectiveness of our complete hierarchical conditioning mechanism, which utilizes both global and part-specific local features.

The \textbf{w/ Global only} variant, which removes the part-specific features and relies solely on the global $C_{global}$ context, shows a degradation in performance. This indicates that while the global feature is crucial for capturing the holistic shape, it lacks the fine-grained guidance necessary for high-fidelity local detail. Without the specific $\{C_{part_{i}}\}$ features, the model struggles to precisely reconstruct the geometry of individual parts, resulting in lower overall accuracy.

\begin{figure}[t]
    \centering
    \includegraphics[width=0.45\textwidth]{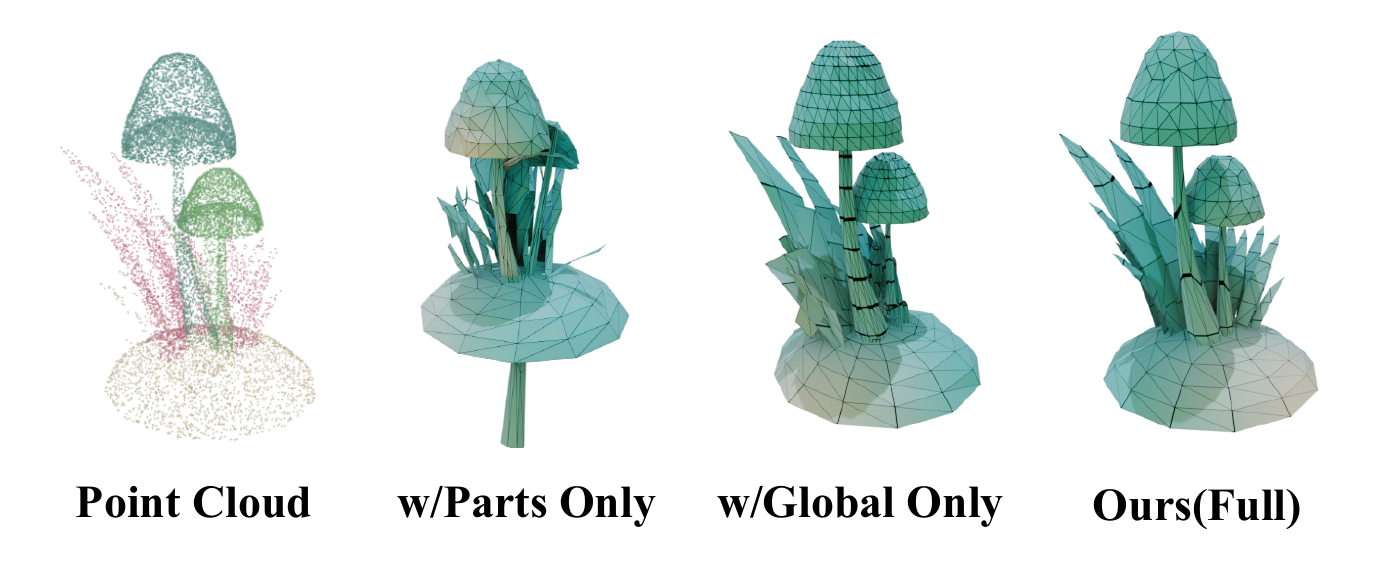}
    \caption{An example of the ablation study.}
    \label{fig:ablation}
\end{figure}

Conversely, the \textbf{w/ Parts only} variant performs the worst by a substantial margin. The poor results strongly suggest that the global context is critical for ensuring structural integrity and correctly assembling the individual parts. An example is given in Figure \ref{fig:ablation}. Even with detailed local information, the model struggles to maintain global coherence, confirming that both $C_{global}$ and $\{C_{part_{i}}\}$ are essential and synergistic components of our design.

\subsection{Efficiency Analysis}

Following the sampling strategy of LLaDA \cite{nie2025llada}, we investigate the potential for parallel generation—a key advantage of discrete diffusion over AR models. While our primary experiments utilize a high-fidelity strategy where the sampling budget $T$ per block is initially set to align with the block length, we further investigate the efficiency potential of our framework.

\begin{table}[h]
    \centering
    \caption{Ablation study on Acceleration Factor $k$. We evaluate the impact of increasing $k$ on the Objaverse dataset. $T$ denotes the sampling steps for each block. Time represents the average inference latency per mesh in seconds.}
    \label{tab:efficiency_ablation}
    \setlength{\tabcolsep}{4pt}
    \begin{tabular}{cc|c|cc}
    \toprule
    Accel. ($k$) & Steps ($T$) & Time (s) $\downarrow$ & CD $\times 10^3$ $\downarrow$ & HD $\downarrow$  \\
    \midrule
    1 (Default) & 1024 & 63.8 & \textbf{17.813} & \textbf{0.238}  \\
    2           & 512  & 33.7 & 27.657 & 0.299 \\
    4           & 256  & \textbf{17.1} & 44.720 & 0.321 \\
    \bottomrule
    \end{tabular}
\end{table}

\begin{figure}[h]
    \centering
    \includegraphics[width=0.45\textwidth]{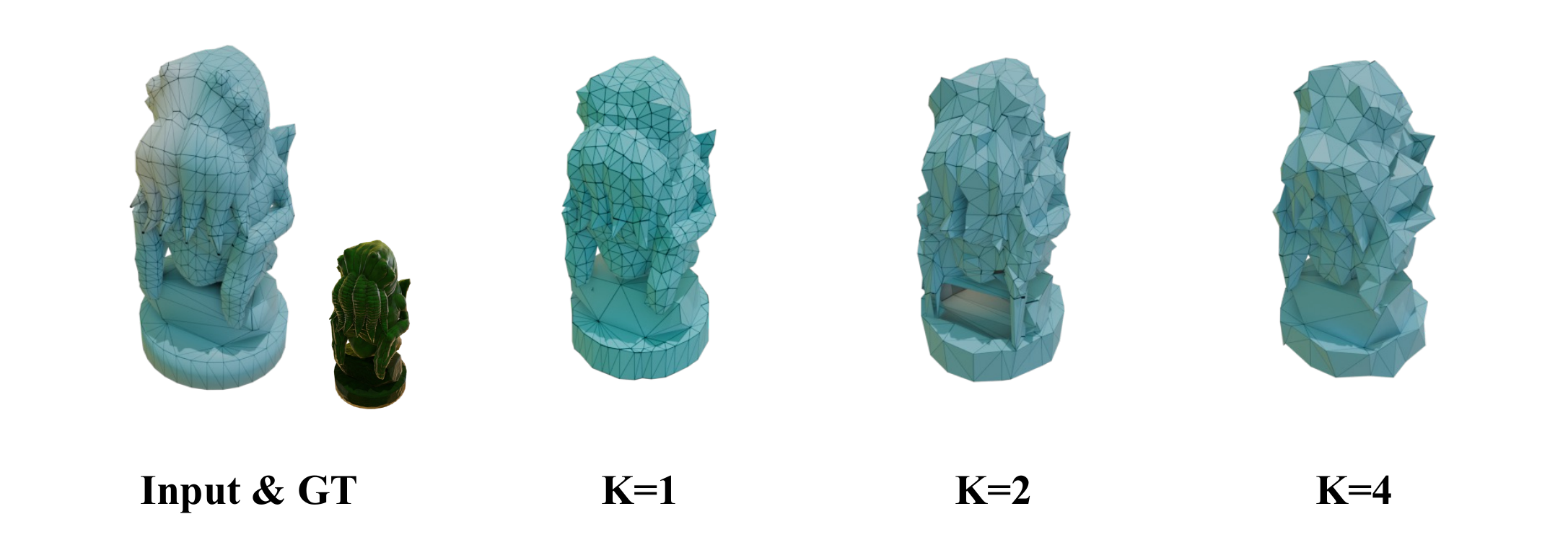}
    \caption{Visual comparison of varying Acceleration Factor $k$.}
    \label{fig:ef_ablation}
    \vspace{-12pt}
\end{figure}

Table~\ref{tab:efficiency_ablation} quantitatively illustrates the trade-off between inference efficiency and generation quality. We introduce $k$ as the \textbf{sampling acceleration factor}, which controls the density of token transition in each step by proportionally reducing the diffusion steps $T$ for each block. While the default setting ($k=1, T=1024$) serves as the high-fidelity baseline, increasing $k$ to $4$ reduces the sampling budget to $T=256$. This achieves a substantial $3.7\times$ speedup, reducing the inference time to $17.1$s. However, this aggressive acceleration forces the model to transition more tokens within fewer iterations, often including tokens with lower prediction confidence. As indicated by the increase in CD to $44.720$, this trade-off comes at the cost of geometric precision. As visualized in Figure \ref{fig:ef_ablation}, while the global topology remains consistent, increasing $k$ leads to increasingly irregular or erroneous high-frequency surface details, such as irregular patches or fractured surfaces. These results demonstrate that to obtain high-fidelity results, an appropriate sampling budget must be maintained at the expense of inference speed.

\section{Conclusion}

We proposed PartDiffuser, a novel semi-autoregressive diffusion framework that resolves the conflict between global structural coherence and local high-fidelity details in 3D mesh generation. Our method operates in a "part-wise manner: it employs autoregression between semantic parts to ensure global topology and parallel discrete diffusion within each part to reconstruct high-frequency geometry. Experimental results demonstrate that PartDiffuser outperforms state-of-the-art methods, particularly in recovering local details often over-smoothed by purely autoregressive models.

\subsection{Limitations and Future Work}

Our work has several limitations, which also point to clear directions for future research. The current model capacity and context window restrict the overall scale and complexity of the generated 3D assets; thus, a key future direction is to scale up our model and explore strategies for longer context windows. Furthermore, we will explore more parallel sampling strategies and systematically study the relationship between the number of sampling steps and the generated mesh quality, in order to better trade off sampling efficiency and output fidelity. Finally, the model's performance depends on the upstream segmentation, and we aim to explore end-to-end joint training and extend PartDiffuser to accept multimodal inputs.

\section{Acknowledgments}
The work was supported by the National Natural Science Foundation of China (No. 62406298).

{
    \small
    \bibliographystyle{ieeenat_fullname}
    \bibliography{main}
}

\clearpage
\setcounter{page}{1}
\maketitlesupplementary

\setcounter{section}{0} 
\renewcommand{\thesection}{\Alph{section}} 

\section{Dataset Construction}
\label{sec:dataset}

As a supplement to the dataset introduction in the main text, we provide a detailed description of the dataset construction process. We utilize Objaverse \cite{deitke2023objaverse} and 3D-Front \cite{fu20213dfront} as our primary data sources. The data preprocessing pipeline consists of four main steps: data filtering, segmentation configuration, serialization augmentation, and block filtering.

Considering computational costs and training stability, we first filter the raw mesh data, retaining only models with a face count fewer than 4000. The filtered data is randomly divided into training, validation, and test sets in a 9:1:1 ratio.

In the segmentation phase, we employ PartField \cite{liu2025partfield} to process the meshes. PartField requires explicit specifications for the minimum and maximum number of clusters. To adapt to geometric structures of varying complexity, we dynamically determine these bounds based on the mesh resolution. Specifically, for a mesh with $F$ faces, the minimum cluster count $K_{min}$ and the maximum cluster count $K_{max}$ are calculated as:

\begin{equation*}
\begin{aligned}
    K_{min} &= \min\left( \left\lfloor \frac{F \times 0.5}{500} \right\rfloor, 4 \right), \\
    K_{max} &= \min\left( \left\lfloor \frac{F \times 2.0}{500} \right\rfloor, 4 \right)
\end{aligned}
\end{equation*}

where $\lfloor \cdot \rfloor$ denotes the floor operation. These bounds are passed to PartField to ensure that the segmentation granularity aligns with the mesh complexity while remaining within a manageable range for the generative model.

\paragraph{Serialization and Augmentation.}
After obtaining the semantic segmentation results, we construct a part adjacency graph, and then randomly select two different parts as starting nodes to perform Breadth-First Search traversals to perform data augmentation on each mesh sample. Subsequently, we employ BPT \cite{weng2025scaling} to encode the mesh following these orders.

\begin{figure}[h]
    \centering
    \includegraphics[width=0.45\textwidth]{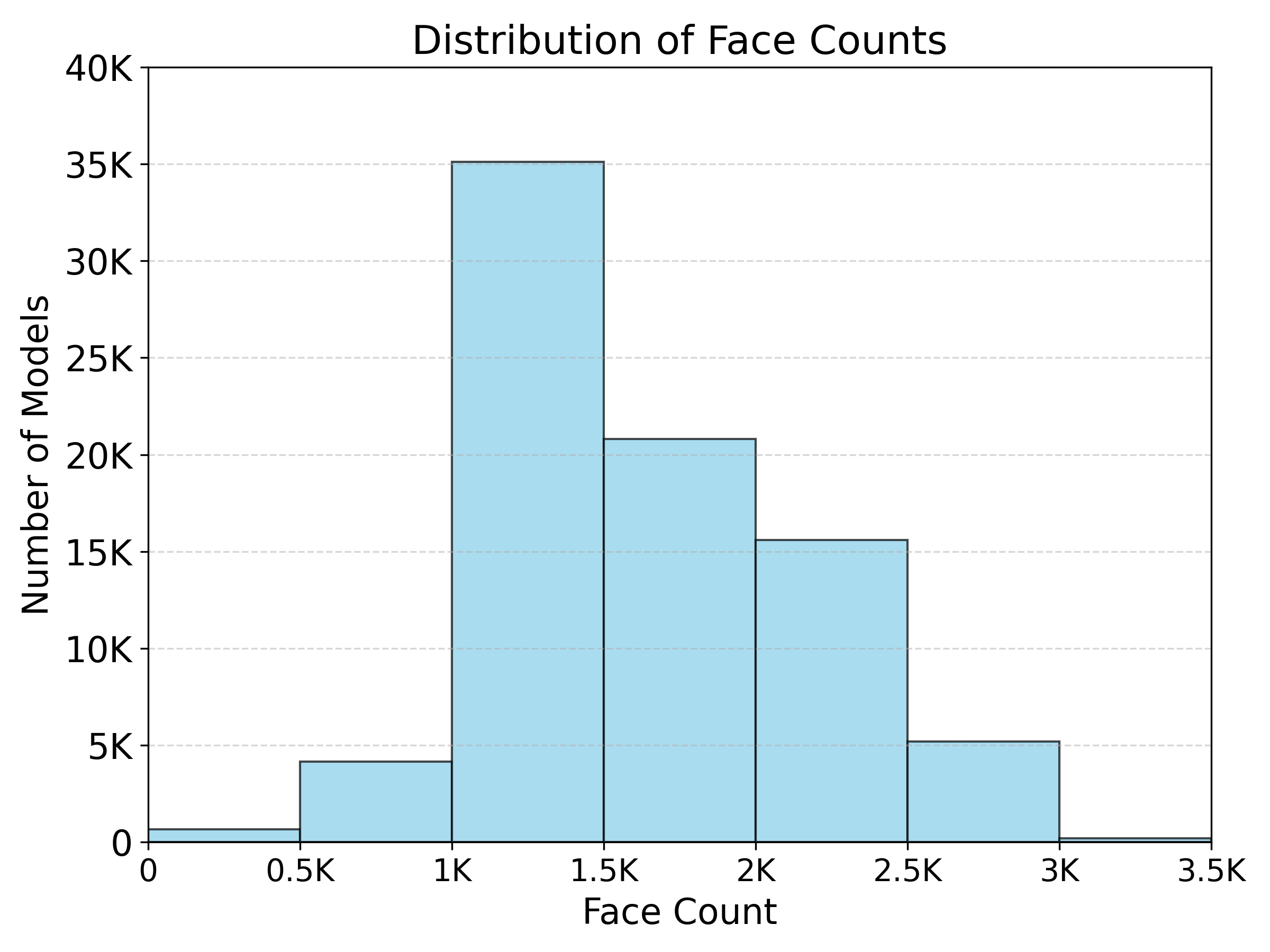}
    \caption{Mesh with different faces in the dataset.}
    \label{fig:face_dist}
\end{figure}

\paragraph{Block Filtering}
To optimize the training efficiency of the discrete diffusion model and prevent an excessive proportion of Pad Tokens in the sequence from causing ineffective computation, we perform a secondary filtering on the encoded sequences. We mandate that the encoded length of each semantic part must fall within the range of 128 to 1024 tokens. For blocks satisfying this condition, we uniformly pad them to a length of 1024 using Pad Tokens. Samples containing blocks that fall outside this range are discarded. The distribution of mesh face counts in the the training set is shown in Figure~\ref{fig:face_dist}.

\section{Training and Sampling Details}
\subsection{Training Strategy}

\begin{figure}[htbp]
    \centering
    \includegraphics[width=0.47\textwidth]{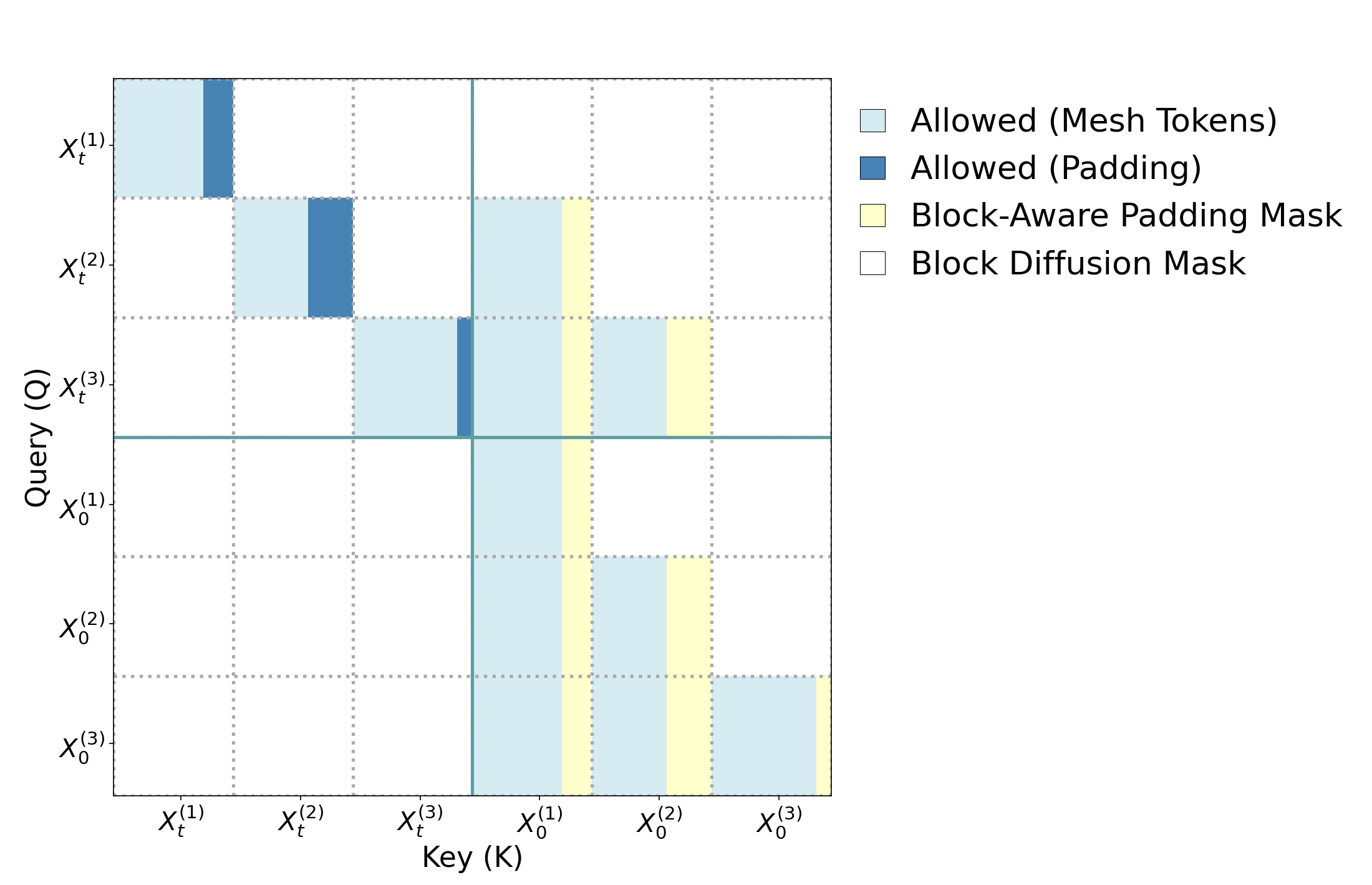}
    \caption{Visualization of the composite attention mask during the parallel training phase, using $N=3$ parts as an example. This mask governs the attention mask used in Section \ref{training strategy}, managing interactions across the $N$ noisy blocks, denoted $X_t$, and the $N$ clean blocks, denoted $X_0$. The legend details the four distinct attention behaviors: standard allowance for mesh tokens, specific allowance for padding tokens, and blockage enforced by either the Block Diffusion Mask or the Block-Aware Padding Mask.}
    \label{fig:train_mask}
\end{figure}

As introduced in the main text, the self-attention module employs a specialized composite mask. This mask combines two distinct components via logical intersection. The Block Diffusion Mask stemming from BD-LM \cite{arriola2025block} enables fine-grained control over the diffusion process by regulating the flow of information between different blocks. The second component, the Block-Aware Padding Mask, works in tight coordination with this hybrid structure. It supports the bidirectional intra-block attention by permitting full visibility within the current denoising block, including padding tokens. Conversely, for the autoregressive inter-block attention, it strictly prevents attention to any padding tokens in other blocks. An example of the composite mask used during the training phase is illustrated in Figure \ref{fig:train_mask}. 

During the training phase, the model operates in a parallel mode, receiving the complete token sequence at once. This sequence is logically divided into $2N$ blocks ($N$ blocks for the noisy version $X_t$ and $N$ blocks for the clean version $X_0$). The composite attention mask is applied to this parallel block structure. This parallelization strategy allows the model to efficiently learn the precise correspondence between token sequences and specific part geometries.

To balance training stability with generative coverage, we adopt a two-stage training curriculum. In the pre-training stage, we adopt the \textit{clipped noise schedule} $\mathcal{U}[\beta, \omega]$ \cite{arriola2025block} to ensure efficient acquisition of core geometric features. In the fine-tuning stage, the model is fine-tuned under a full linear schedule $\mathcal{U}[0, 1]$, enabling it to adapt to the complete denoising trajectory required during inference and improving overall robustness across all noise levels.

\subsection{Sampling Strategy}

\begin{algorithm}[htbp]
\caption{Sampling Strategy}
\label{alg:sampling}
\small 
\begin{algorithmic}[1]
\Require Hierarchical geometric condition $C_{\text{pc}} = [C_{\text{global}}, C_{\text{part}_1}, \dots, C_{\text{part}_N}]$, Diffusion steps $T$, block length $L_{\text{block}}$
\State Initialize accumulated sequence $X_{\text{accum}} \gets \emptyset$
\For{$i \leftarrow 1$ to $N$}
    \State $Z \gets \text{SamplePrior}(B, L_{\text{block}})$ 
    \State $X_{\text{context}} \gets \text{concat}(X_{\text{accum}}, Z)$ 
    \State $C_{\text{dyn}} \gets \text{concat}(C_{\text{global}}, C_{\text{part}_i})$ 
    \For{$t \leftarrow T$ down to $1$} 
        \State // Pass full context, focus on last block
        \State $P(X_0 | X_t) \gets \text{Model}(X_{\text{context}}, t, C_{\text{pc}}, \text{block\_idx}=i)$
    
    \State // CrossAttn uses $C_{\text{dyn}}$ due to block\_idx
        \State $Z_{\text{denoised}} \gets \text{SampleStep}(P(X_0), X_{\text{context}}, t, t-1)$
        \State $X_{\text{context}} \gets \text{update\_context}(X_{\text{accum}}, Z_{\text{denoised}})$
    \EndFor
    \State $X_{\text{accum}} \gets \text{concat}(X_{\text{accum}}, Z_{\text{denoised}})$
\EndFor
\State \textbf{return} $X_{\text{accum}}$
\end{algorithmic}
\end{algorithm}

Algorithm \ref{alg:sampling} provides the pseudo-code for our semi-autoregressive sampling process. The generation reconstructs one semantic part at a time, consisting of $N$ strides corresponding to the total number of parts determined by the input condition $C_{\text{pc}}$.

\section{Sampling Analysis}
\label{sec:sampling}

\subsection{Robustness Analysis}

To analyze robustness, we conducted a brief experiment using SAMPart3D \cite{yang2024sampart3d} as an alternative segmentation upstream. 

\begin{figure}[t]
    \centering
    \includegraphics[width=0.45\textwidth]{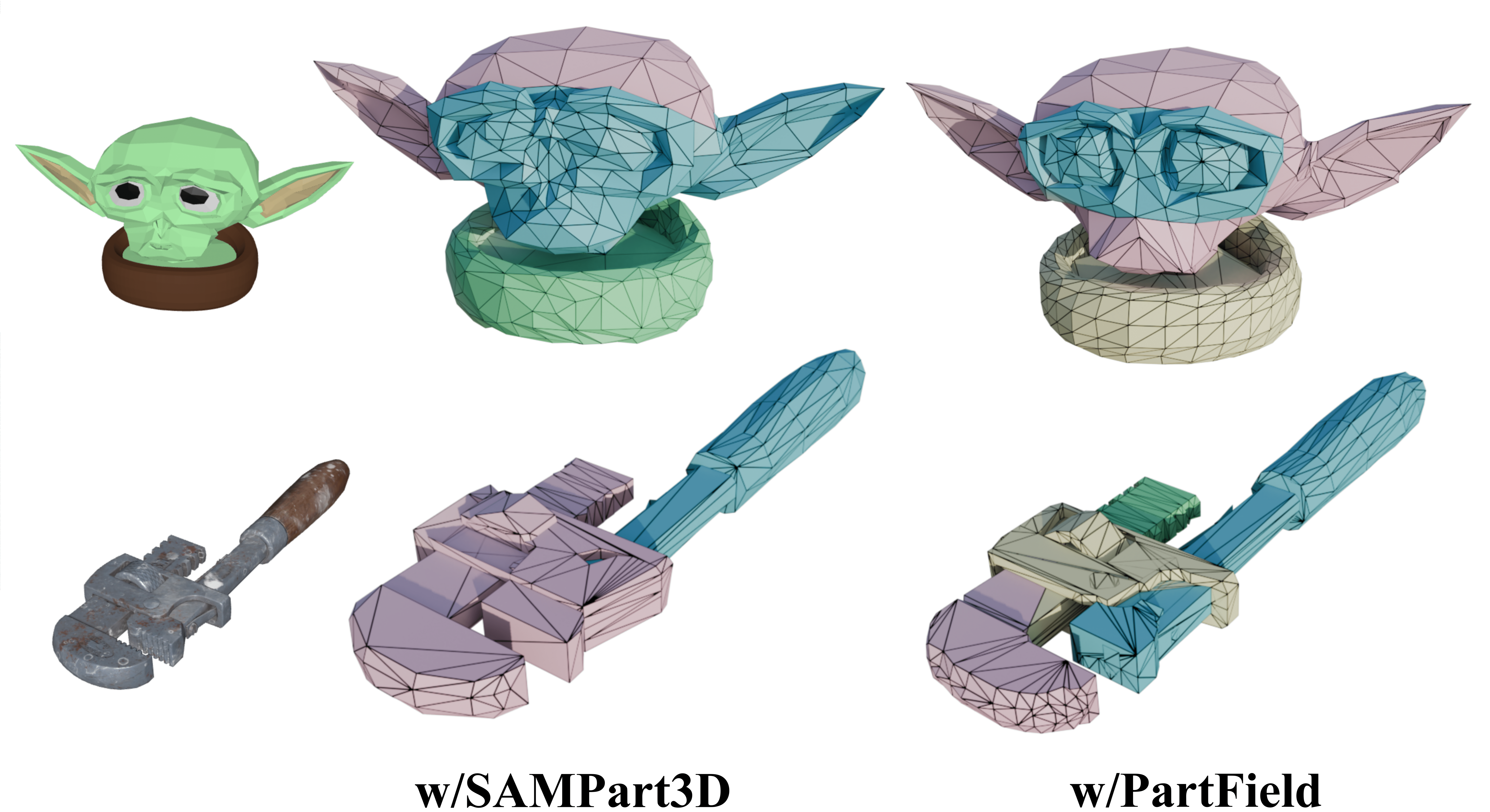}
    \label{fig:robust}
    \caption{Experiment on robustness to different segmentation strategies.}
\end{figure}
Quantitatively, using SAMPart3D on Objaverse yields a CD of $19.11 \times 10^{-3}$, closely matching the original $17.81$ using PartField.
The analysis demonstrates that our model is reasonably robust to different segmentation strategies. The denoising process prioritizes the completeness and smoothness of local point cloud geometry rather than strict semantic boundaries. However, as shown in the comparison, different segmentation strategies can affect local detail representation. 
Coarser segments may reach this capacity sooner, leading to relative simplification, while more balanced partitions utilize the token budget effectively for higher fidelity.

\subsection{Sampling Process}

\begin{figure*}[h]
    \centering
    \includegraphics[width=0.9\textwidth]{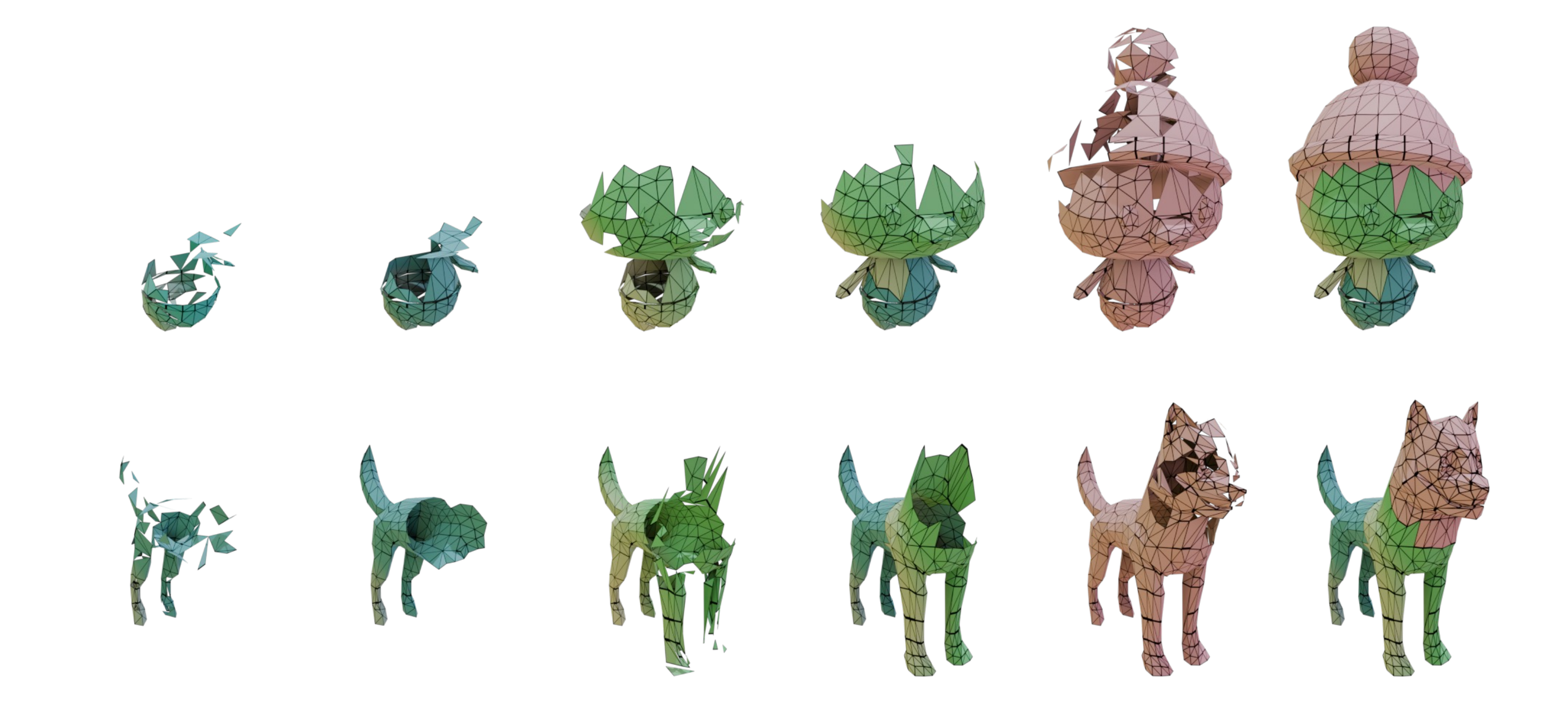}
    \caption{Visualization of the part-wise sampling process. The process evolves from left to right: (1) The first part being denoised, (2) The first part completed, (3) The second part being denoised, (4) The second part completed, and so on. This highlights how PartDiffuser combines part-level autoregression with discrete diffusion.}
    \label{fig:sampling}
    \vspace{-12pt}
\end{figure*}

To intuitively illustrate the core mechanism of PartDiffuser, we visualize the part-by-part sampling process in Figure~\ref{fig:sampling}. Crucially, the visualization highlights a distinct behavior compared to conventional fully AR approaches. In AR methods, mesh faces are typically generated sequentially along a fixed topological trajectory or backbone. In contrast, the intra-part diffusion process exhibits a parallel generation pattern: faces do not propagate linearly from a seed point. Instead, valid geometry materializes simultaneously across disparate spatial locations within the part's volume. As observed in the intermediate steps, the model does not merely extend a continuous partial surface; rather, it often constructs a sparse structural "framework" or "skeleton" of the part first. This aligns with the confidence-based nature of discrete diffusion, where the model prioritizes high-confidence structural tokens. The model's ability to establish the overall geometric scaffolding may help mitigate the risk of local over-complexity.

Besides, the visualization also verifies semi-autoregressive design, where subsequent parts are explicitly conditioned on the completed geometry of their predecessors. It can be observed that in the noisy stages, the newly emerging faces tend to align with the boundaries of the previously generated fixed parts. This may demonstrate that the model utilizes the context from previous parts to ensure topological connectivity.

\end{document}